\documentclass[
]{ceurart}

\usepackage{listings}
\usepackage{makecell}
\usepackage{tabularx}
\usepackage{eso-pic}
\usepackage{xcolor, graphicx}
\usepackage{transparent}
\usepackage{xkeyval}

\begin{document}

\AddToShipoutPictureBG*{%
  \AtPageLowerLeft{%
    \put(10,214){%
      \rotatebox{90}{\textcolor[rgb]{0.55,0.55,0.902}{\ttfamily CEUR-WS.org/Vol-3740/paper-09.pdf}}%
    }%
  }%
}

\AddToShipoutPictureBG*{%
  \AtPageLowerLeft{%
    \raisebox{13.7mm}{
    \hspace{3mm}
    \transparent{0.53}
    \includegraphics[width=26mm]{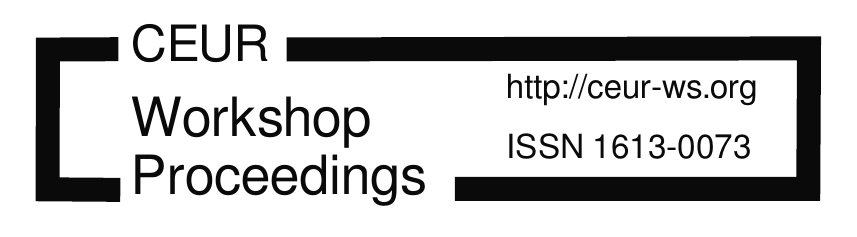}%
    }%
  }%
}

\copyrightyear{2024}
\copyrightclause{Copyright for this paper by its authors.
  Use permitted under Creative Commons License Attribution 4.0
  International (CC BY 4.0).}

\conference{CLEF 2024: Conference and Labs of the Evaluation Forum, September 09–12, 2024, Grenoble, France}

\title{Multilingual Clinical NER for Diseases and Medications Recognition in Cardiology Texts using BERT Embeddings}

\tnotetext[0]{Disclaimer: The concepts and information presented in this paper are based on research results that are not commercially available.}

\author[1,2]{Manuela Daniela Danu}[email=manuela.voinea@siemens.com,]
\cormark[1]
\author[1]{George Marica}[email=george.marica@siemens.com,]
\author[1, 2]{Constantin Suciu}[email=constantin.suciu@siemens.com,]
\author[1,2]{Lucian Mihai Itu}[email=lucian.itu@siemens.com,]
\author[3]{Oladimeji Farri}[email=oladimeji.farri@siemens-healthineers.com,]

\address[1]{Advanta, Siemens SRL, 15 Noiembrie Bvd, 500097 Brasov, Romania}
\address[2]{Automation and Information Technology, Transilvania University of Brasov, 5 Mihai Viteazul Street, 500174 Brasov, Romania}
\address[3]{Digital Technology and Innovation, Siemens Healthineers, 755 College Rd E, 08540 Princeton, NJ, United States}

\cortext[1]{Corresponding author.}

\begin{abstract}
  The rapidly increasing volume of electronic health record (EHR) data underscores a pressing need to unlock biomedical knowledge from unstructured clinical texts to support advancements in data-driven clinical systems, including patient diagnosis, disease progression monitoring, treatment effects assessment, prediction of future clinical events, etc. While contextualized language models have demonstrated impressive performance improvements for named entity recognition (NER) systems in English corpora, there remains a scarcity of research focused on clinical texts in low-resource languages. To bridge this gap, our study aims to develop multiple deep contextual embedding models to enhance clinical NER in the cardiology domain, as part of the BioASQ MultiCardioNER shared task. We explore the effectiveness of different monolingual and multilingual BERT-based models, trained on general domain text, for extracting disease and medication mentions from clinical case reports written in English, Spanish, and Italian. We achieved an F1-score of 77.88\% on Spanish Diseases Recognition (SDR), 92.09\% on Spanish Medications Recognition (SMR), 91.74\% on English Medications Recognition (EMR), and 88.9\% on Italian Medications Recognition (IMR). These results outperform the mean and median F1 scores in the test leaderboard across all subtasks, with the mean/median values being: 69.61\%/75.66\% for SDR, 81.22\%/90.18\% for SMR, 89.2\%/88.96\% for EMR, and 82.8\%/87.76\% for IMR.
\end{abstract}


\maketitle

\section{Introduction}

With the increasing amount of available electronic health record (EHR) data, clinical natural language processing (NLP) tasks have become significantly important for extracting valuable information from unstructured clinical texts \cite{rubel2020biobertpt}. Named Entity Recognition (NER) is a key NLP task used to identify meaningful entities within these texts, such as anatomical structures, diseases and disorders, signs and symptoms, procedures, and medications \cite{rubel2020biobertpt, kundeti2016clinical}. Consequently, this facilitates various data analysis applications, ranging from predicting future clinical events \cite{jin2018improving} to summarization \cite{riccio2023healthcare} and relation extraction between entities (e.g., drug-to-drug interactions \cite{zaikis2020drug}, symptom-disease relationship \cite{abulaish2019disease}, patient-procedure association \cite{rink2011automatic}, etc.) 

Despite recent advances in deep learning methods for NER \cite{lee2020biobert, kanakarajan2021bioelectra}, extracting structured information from the vast amounts of unstructured and often noisy clinical documents in EHR systems remains challenging due to the highly specialized medical language, which varies considerably across different medical specialties, as well as due to the prevalence of misspellings, abbreviations, and use of synonyms to express clinical concepts \cite{rubel2020biobertpt}.

While contextualized language models have recently improved the performance of NER systems for English corpora \cite{lee2020biobert, alsentzer2019publicly, li2019fine}, there is a notable lack of research focused on clinical texts in low-resource languages. To address this gap, our study aims to develop multiple deep contextual embedding models for English, Spanish, and Italian to enhance clinical NER in the cardiology domain, as part of the MultiCardioNER shared task \cite{multicardioner2024overview, nentidis2024bioasq, BioASQ2024overview}. The MultiCardioNER task is part of the twelfth edition of the large-scale biomedical semantic indexing and question answering challenge (BioASQ) \cite{nentidis2024bioasq, BioASQ2024overview}, a long-standing initiative aiming to advance research by developing methods and tools that leverage the vast amount of online information to meet the needs of biomedical researchers and practitioners. This initiative seeks to provide efficient and rapid access to the continuously expanding resources and knowledge in the biomedical field. 

MultiCardioNER \cite{multicardioner2024overview, nentidis2024bioasq, BioASQ2024overview} is a shared task that aims to automatically identify two key clinical concepts in medical documents pertaining to cardiology, namely diseases and medications. This task focuses on adapting clinical NER systems to effectively work across multiple languages - primarily Spanish, English, and Italian - for two different subtasks: (1) diseases recognition in Spanish cardiology texts, and (2) medications recognition in cardiology texts written in Spanish, English, and Italian. Both subtasks involve reading and analyzing clinical texts to identify the clinical entities mentioned in the text and using the BRAT format to mark the starting and ending positions of these entities.

In this paper, we created four different monolingual models: (1) Spanish Diseases Recognition (SDR), (2) Spanish Medications Recognition (SMR), (3) English Medications Recognition (EMR), and (4) Italian Medications Recognition (IMR). Additionally, we developed two multilingual models: one specialized for Spanish Diseases Recognition (Multi-SDR) and another for Medications Recognition across all three targeted languages (Multi-MMR). We applied transfer learning techniques by fine-tuning BERT-based \cite{devlin2018bert} contextual embeddings, originally trained on general domain text in each of the three languages, for the biomedical domain to extract diseases and medications from clinical reports.

\section{Related Work}

In clinical and biomedical NER, recent studies have explored various methodologies to enhance performance. A key model in this domain is multilingual BERT (M-BERT) \cite{devlin2018bert}, trained on 104 Wikipedia languages, which excels in various tasks without explicit cross-lingual alignment \cite{pires-etal-2019-multilingual}, outperforming models based on cross-lingual embeddings \cite{wu2019beto}.

\cite{tian2020improving} improved biomedical NER by incorporating syntactic information, enhancing recognition of complex entity relationships (ORCID). \cite{perez2021identifying} focused on de-identifying Spanish medical texts via NER and entity randomization, achieving high recall rates on radiology reports and MEDDOCAN \cite{marimon_2020_4279323} challenge data. \cite{kanakarajan2021bioelectra} developed BioELECTRA, a biomedical text encoder using discriminators, which outperformed several baselines on multiple biomedical NER benchmarks by leveraging ELECTRA's efficiency and accuracy in text encoding.

\cite{kocaman2021biomedical} developed a scalable NER system for large biomedical datasets, emphasizing real-time processing and high accuracy. \cite{carrino2022pretrained} focused on pre-trained biomedical language models for clinical NLP in Spanish, addressing the need for multilingual capabilities in biomedical NER. \cite{zhang2022optimizing} optimized a bi-encoder for NER using contrastive learning, introducing dynamic thresholding to improve accuracy, especially for nested entities, with significant gains on datasets like ACE \cite{walker2005ace} and GENIA \cite{10.5555/1289189.1289260}. \cite{khandelwal2022biomedical} used a novel schema with distant supervision to enhance NER accuracy, showing that domain-specific schema can supplement limited annotated data effectively.

\cite{hu2023zero} used ChatGPT \cite{chatgpt} for zero-shot clinical entity recognition with prompt engineering, showing it outperforms GPT-3 \cite{brown2020language} but trails behind fine-tuned BioClinicalBERT \cite{alsentzer2019publicly} models. \cite{bhattacharya2023improving} leveraged transfer learning and asymmetric tri-training, combining labeled and pseudo-labeled data to boost NER performance across biomedical datasets.

To advance the development of medical NER systems, the BioASQ challenge proposed multiple clinical NER tasks to be solved over time, such as automatic detection and normalization of disease mentions from clinical texts (DisTEMIST) \cite{miranda2022overview} or medical procedure detection and entity linking (MedProcNER) \cite{lima2023overview}. Most participating teams employed Transformer-based and large language models in their approaches.

\section{Methods}
\subsection{Datasets}
With a focus on adapting general medical NER systems for diseases and medications across multiple languages, the MultiCardioNER \cite{multicardioner2024overview, lima_lopez_2024_11368861} task leverages several datasets. Specifically, it utilizes a training collection of 1000 general clinical case reports in Spanish, covering various medical specialties such as oncology, urology, ophthalmology, dentistry, pediatrics, primary care, allergology, radiology, psychiatry, and more \cite{lima_lopez_2024_11368861, miranda2022overview}. These reports were annotated with diseases and medications, resulting in two distinct corpora, namely DisTEMIST \cite{lima_lopez_2024_11368861, miranda2022overview, miranda_escalada_2023_7614764} and DrugTEMIST \cite{lima_lopez_2024_11368861}. The DrugTEMIST \cite{lima_lopez_2024_11368861} corpus was also released in English and Italian. Since the original 1000 clinical case reports belong to the Spanish Clinical Case Corpus (SPACCC) \cite{ander_intxaurrondo_2019_2560316}, the multilingual DrugTEMIST \cite{lima_lopez_2024_11368861} dataset was originally created in Spanish and then transferred into English and Italian using machine translation and lexical annotation projection. The result of this process was revised and validated by clinical experts who are native speakers of each language.

For the domain adaptation part of the task, MultiCardioNER \cite{lima_lopez_2024_11368861} leverages a collection of 508 annotated cardiology clinical case reports (CardioCCC), divided into 258 for development and 250 for testing. The annotation process followed the same guidelines as the DisTEMIST \cite{farre_maduell_2022_6477407} and DrugTEMIST \cite{lima_lopez_2024_11065433} corpora, with the medication part also released in Spanish, English and Italian. In addition to the test set, an auxiliary collection of multilingual clinical case reports, referred to as the background set, is provided to facilitate the creation of a silver standard corpus and ensure the developed systems can effectively scale up to larger content collections.

All datasets were manually annotated by clinical experts using the BRAT annotation tool \cite{stenetorp2012brat}, following well-defined annotation guidelines \cite{farre_maduell_2022_6477407, lima_lopez_2024_11065433} defined after several cycles of quality control and annotation consistency analysis.

\subsection{Experiments}
In this work, we treated the automatic named entity recognition (NER) of diseases and medications in clinical case reports as a multi-label token classification task. To accomplish this, we employed pre-existing BERT models \cite{devlin2018bert} for NER in the general domain for each of the three languages (Spanish, English, and Italian), as well as a multilingual model, and further fine-tuned them for the biomedical domain using the MultiCardioNER dataset \cite{lima_lopez_2024_11368861}.

We experimented with the following BERT-based models, specifically trained to perform NER:

\begin{itemize}
\item \textbf{bert-spanish-cased-finetuned-ner} \cite{hf-spanish}: a Spanish BERT cased model based on BETO \cite{CaneteCFP2020}. Originally fine-tuned on the Spanish dataset of the CoNLL-2002 Shared Task \cite{tjong-kim-sang-2002-introduction}, BETO was further fine-tuned on the Catalan and Basque subsets of the CoNLL-2007 dataset \cite{nivre2007conll}, resulting in the bert-spanish-cased-finetuned-ner model, which focuses on recognizing persons (PER), organizations (ORG), locations (LOC), and miscellaneous (MISC) entities within Spanish text documents.

\item \textbf{bert-base-NER} \cite{hf-english}:
a BERT cased model fine-tuned on the English version of the standard CoNLL-2003 dataset \cite{tjong-kim-sang-de-meulder-2003-introduction}. It was trained to recognize four types of entities, namely locations (LOC), organizations (ORG), persons (PER), and miscellaneous (MISC).

\item \textbf{bert-italian-finetuned-ner} \cite{hf-italian}: an Italian BERT cased model fine-tuned on the WikiANN dataset \cite{pan-etal-2017-cross}, which consists of Wikipedia articles annotated with LOC (location), PER (person), and ORG (organisation) tags.

\item \textbf{bert-base-multilingual-cased-ner-hrl} \cite{hf-multilingual}: a named entity recognition model for 10 high-resourced languages (Arabic, German, English, Spanish, French, Italian, Latvian, Dutch, Portuguese and Chinese) based on a fine-tuned multilingual cased BERT model. It has been trained to recognize three types of entities: locations (LOC), organizations (ORG), and persons (PER).
\end{itemize}

All these BERT-based models utilize the standard Beginning-Inside-Outside (BIO) format \cite{ramshaw1999text} for tagging entities. This format is crucial as it allows NER to be approached as a multi-label classification task, where words are labelled B if they represent the beginning of an entity, I if they are inside an entity, and O if they are outside any entity. This labeling method effectively distinguishes between the beginning and continuation of an entity, thereby simplifying the task of identifying entity boundaries.

\begin{figure}
  \centering
  \includegraphics[width=\linewidth]{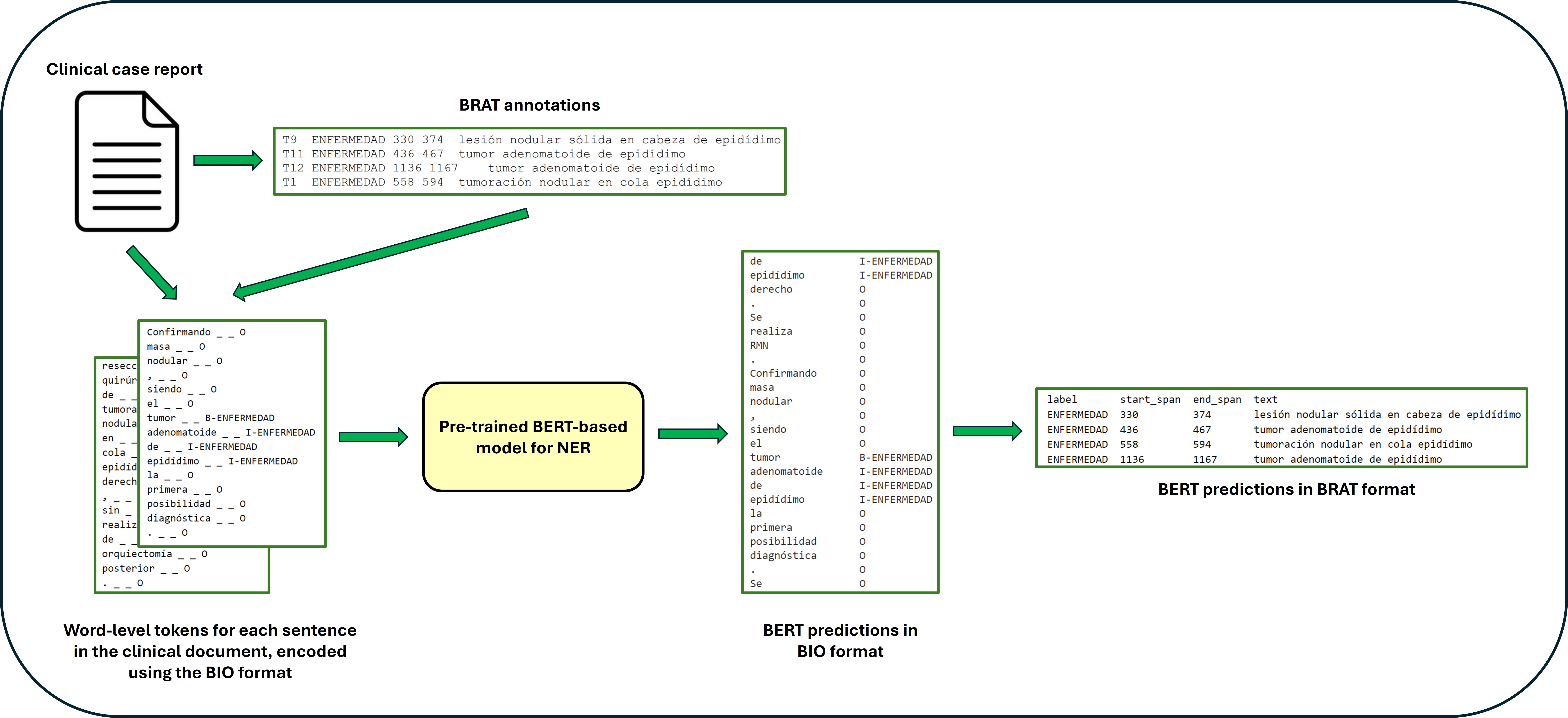}
  \caption{Overview of the prediction pipeline. As a pre-processing step, the clinical case reports are split into sentences, and further segmented into word-level tokens. By leveraging the available BRAT annotations, the word-level tokens are encoded using the BIO format and used to fine-tune BERT-based models on the MultiCardioNER dataset. The output from the BERT models, also in BIO format, is then post-processed to comply with BRAT format.}
  \label{pipeline}
\end{figure}

Before performing the clinical domain adaptation of the general domain BERT-based models, the medical reports undergo a pre-processing step which involves splitting them into sentences to ensure a sequence length of less than or equal to 256. These sentences are then further segmented into word-level tokens while preserving their start and end offsets with respect to the original report. The word-level tokens are encoded in BIO format and used to fine-tune BERT-based models on the MultiCardioNER dataset. The output from the BERT models is then post-processed to comply with BRAT format. Figure~\ref{pipeline} provides an overview of the prediction pipeline.

Details regarding the label lists used for each subtask, as well as the hyper-parameters configuration employed in the experiments, are provided in sections 3.2.1 and 3.2.2, respectively.

\subsubsection{Subtask 1: Diseases Recognition in Spanish Cardiology Texts}

For the subtask aiming to address the recognition of diseases in Spanish cardiology texts, we leveraged the pre-trained \emph{bert-spanish-cased-finetuned-ner} and \emph{bert-base-multilingual-cased-ner-hrl} models and further fine-tuned them on the MultiCardioNER dataset. Specifically, we employed the DisTEMIST corpora as the training set for the general clinical domain adaptation part of the task and used the disease-annotated version of the Spanish CardioCCC clinical cases as the development set to identify the best-performing models in the cardiology domain, resulting in the Clinical-SDR and MultiClinical-SDR models. We additionally experimented with fine-tuning these models on the CardioCCC development set, leading to the creation of the cardiology-specialized Cardio-SDR and MultiCardio-SDR models.

Following the standard BIO format \cite{ramshaw1999text}, we defined our label list as follows: B-ENFERMEDAD, I-ENFERMEDAD, O, [CLS], and [SEP]. B-ENFERMEDAD and I-ENFERMEDAD denote the beginning and continuation of disease mentions within text sequences, whereas the label O corresponds to word-level tokens outside any recognized entity. Additionally, the [CLS] token indicates the commencement of a sentence, while the [SEP] token marks its termination. Notably, the [CLS] token also serves as a placeholder for the [PAD] token within the text sequences. 

The Spanish Diseases Recognition (SDR) models were fine-tuned on an NVIDIA GeForce RTX 3090 (24GB) GPU for 10 epochs. The utilized hyper-parameters configuration includes a maximum sequence length of 256, a batch size of 8, and a learning rate of $9e^{-6}$. Predictions were generated for both the test and background sets, but the evaluation exclusively considered the predictions achieved on the test set. In addition to the test set results, we also reported the development set results to identify any discrepancies between them and, hence, detect potential overfitting or any issues related to the data split distributions.

\subsubsection{Subtask 2: Multilingual (Spanish, English and Italian) Medications Recognition in Cardiology Texts}

For the second subtask, which focuses on the recognition of medications in cardiology texts written in Spanish, English, and Italian, we employed three monolingual pre-trained models (\emph{bert-spanish-cased-finetuned-ner}, \emph{bert-base-NER}, and \emph{bert-italian-finetuned-ner}), each specialized for one of the three languages, as well as a multilingual model (\emph{bert-base-multilingual-cased-ner-hrl}), and subsequently fine-tuned them on the MultiCardioNER dataset. Therefore, we leveraged the DrugTEMIST corpora in each of the three languages as the training sets for the general clinical domain adaptation part of the task and used the medication-annotated version of the CardioCCC clinical cases in Spanish, English, and Italian as the development sets to identify the best performing models in the cardiology domain, resulting in the Clinical-SMR, Clinical-EMR, Clinical-IMR, and MultiClinical-MMR models.  We again conducted additional experiments by fine-tuning these models on the CardioCCC development sets, thereby achieving the cardiology-specialized Cardio-SMR, Cardio-EMR, Cardio-IMR, and MultiCardio-MMR models. It is worth noting that the multilingual model was trained on an aggregated dataset encompassing all three languages, but separately evaluated for each language to assess its performance across different linguistic contexts.  

In accordance with the standard BIO format \cite{ramshaw1999text}, we defined the label list for this subtask as B-FARMACO, I-FARMACO, O, [CLS], and [SEP]. The tags B-FARMACO and I-FARMACO denote the beginning and continuation of medication mentions within text sequences, while the label O marks the word-level tokens not associated with any recognized entity. Additionally, as in Subtask 1, the [CLS] token indicates the beginning of a sentence, while the [SEP] token marks its end. In this context, the [CLS] token also serves as a placeholder for the [PAD] token within the text sequences. 

The Spanish Medications Recognition (SMR), English Medications Recognition
(EMR), Italian Medications Recognition (IMR), and Multilingual Medications Recognition (MMR) models were independently fine-tuned on an NVIDIA GeForce RTX 3090 (24GB) GPU for 10 epochs. The utilized hyper-parameters configuration is identical to that employed for Subtask 1 and consists of a maximum sequence length of 256, a batch size of 8, and a learning rate of $9e^{-6}$. Predictions were generated for both the test and background sets. However, the evaluation exclusively considered the predictions obtained on the test set. In addition to the test set results, we also reported the development set results to identify any mismatch between them, which could indicate overfitting or issues related to the data split distributions.

\section{Results}
\begin{table*}[t]
  \caption{Evaluation results on the development and test sets for the MultiCardioNER task. The best results on the test sets are highlighted in bold. The experiments marked with an (*) were conducted after the MultiCardioNER evaluation period and are not included in the official leaderboard.}
  \label{tab:eval_results}
  \setlength\tabcolsep{3.95pt}
  \begin{tabularx}{\linewidth}{llccccccc}
    \toprule
    Subtask & Model & Fine-tuning & \makecell{Dev \\ Precision} & \makecell{Dev \\ Recall} & \makecell{Dev \\ F1-score} & \makecell{Test \\ Precision} & \makecell{Test \\ Recall} & \makecell{Test \\ F1-score} \\
    \midrule
    Track1 (ES) & Clinical-SDR & No & 0.6674 & 0.6243 & 0.6451 & 0.6758 & 0.6437 & 0.6593 \\
    Track1 (ES) * & Cardio-SDR & Yes & 0.9713 & 0.9535 & 0.9623 & \textbf{0.7739} & \textbf{0.7837} & \textbf{0.7788}\\
    Track1 (ES) * & MultiClinical-SDR & No & 0.6355 & 0.6118 & 0.6234 & 0.6387 & 0.6268 & 0.6327\\
    Track1 (ES) * & MultiCardio-SDR & Yes & 0.9406 & 0.9360 & 0.9383 & 0.7717 & 0.7788 & 0.7753\\
    \midrule
    
    Track2 (ES) & Clinical-SMR & No & 0.9019 & 0.8753 & 0.8884 & 0.8928 & 0.8778 & 0.8852\\
    Track2 (ES) * & Cardio-SMR & Yes & 0.9804 & 0.9562 & 0.9681 & 0.9289 & 0.9045 & 0.9165\\
    Track2 (ES) * & MultiClinical-MMR & No & 0.8783 & 0.8681 & 0.8732 & 0.8974 & 0.8807 & 0.8890\\
    Track2 (ES) * & MultiCardio-MMR & Yes & 0.9790 & 0.9482 & 0.9634 & \textbf{0.9341} & \textbf{0.9080} & \textbf{0.9209}\\
    \midrule
    
    Track2 (EN) & Clinical-EMR & No & 0.8866 & 0.8625 & 0.8744 & 0.8685 & 0.8791 & 0.8738\\
    Track2 (EN) * & Cardio-EMR & Yes & 0.9575 & 0.9155 & 0.9360 & \textbf{0.9277} & 0.9018 & 0.9146\\
    Track2 (EN) * & MultiClinical-MMR & No & 0.8833 & 0.8594 & 0.8712 & 0.8920 & 0.8826 & 0.8873\\
    Track2 (EN) * & MultiCardio-MMR & Yes & 0.9681 & 0.9550 & 0.9615 & 0.9121 & \textbf{0.9227} & \textbf{0.9174}\\
    \midrule
    
    Track2 (IT) & Clinical-IMR & No & 0.9122 & 0.8801 & 0.8958 & 0.8891 & 0.8689 & 0.8789\\
    Track2 (IT) * & Cardio-IMR & Yes & 0.9518 & 0.9250 & 0.9382 & 0.8994 & \textbf{0.8789} & \textbf{0.8890}\\
    Track2 (IT) * & MultiClinical-MMR & No & 0.8868 & 0.8603 & 0.8734 & 0.8747 & 0.8378 & 0.8558\\
    Track2 (IT) * & MultiCardio-MMR & Yes & 0.9772 & 0.9455 & 0.9611 & \textbf{0.9046} & 0.8694 & 0.8867\\
  \bottomrule
\end{tabularx}
\end{table*}

In this work, we evaluated the developed systems using a flat evaluation approach \cite{kosmopoulos2015evaluation} by comparing the automatically generated results with those obtained by domain experts through manual annotation. The primary focus was on identifying and classifying clinical mentions of diseases and medications in cardiology reports. The performance metrics employed for flat evaluation include micro-averaged precision, recall, and F1-score (MiF). These metrics were computed based on the exact matches of the predicted entities and the annotated ground-truth. Table~\ref{tab:eval_results} summarises the evaluation results obtained on the development and test sets using the official-released evaluation library for the MultiCardioNER task. In the test set evaluation, we achieved the following F1-scores: 77.88\% for Spanish Diseases Recognition (SDR), 92.09\% for Spanish Medications Recognition (SMR), 91.74\% for English Medications Recognition (EMR), and 88.9\% for Italian Medications Recognition (IMR).

These results surpass the mean and median F1 scores in the test leaderboard across all subtasks, with the mean/median values being: 69.61\%/75.66\% for SDR, 81.22\%/90.18\% for SMR, 89.2\%/88.96\% for EMR, and 82.8\%/87.76\% for IMR.

The experiments marked with an (*) in Table~\ref{tab:eval_results} were conducted after the MultiCardioNER evaluation period and are not included in the official leaderboard. However, these supplementary experiments provide further insights beyond the primary evaluation results. For instance, the fine-tuning process considerably enhances performance across all developed systems. Additionally, employing a multilingual model proves beneficial in certain substasks, such as Spanish Medications Recognition (SMR) and English Medications Recognition (EMR), resulting in an improved F1-score from 91.65\% (achieved by the subsequent best performing model) to 92.09\%, and from 91.46\% to 91.74\%, respectively.

By comparing the results from the development and test sets, we can assess potential discrepancies between these data splits and identify issues such as overfitting or distributional disparities. These insights are crucial for enhancing model robustness and generalization, which are essential for successfully utilizing the developed systems in real-world clinical scenarios. As illustrated in Table~\ref{tab:eval_results}, non-fine-tuned models exhibit similar evaluation metrics on both the development and test sets. For these models, the development set was solely used to select the best-performing model across different checkpoints. This consistency confirms that the two data splits originate from the same distribution. In contrast, fine-tuned models \textendash\ trained on the development set \textendash\ demonstrate a performance gap between the two sets. While some degree of performance difference is expected due to the model’s exposure to the development data during training, excessively large gaps suggest overfitting. This is the case of Spanish Diseases Recognition (SDR) models, where the performance gap between the development and test sets is 18.35\% for Cardio-SDR and 16.3\% for MultiCardio-SDR. For all other fine-tuned models, the F1-score on the development set is only slightly higher than that computed on the test set, with differences ranging from 1.95\% to 7.44\%. Although these differences may indicate some overfitting, they do not reach a severe extent. One plausible explanation for overfitting in these cases could be that the model is too complex for the limited diversity of cardiology-specific entities present in the development set. As a result, the model may capture specific patterns from the training data but struggle to generalize to new data. 

\begin{figure}
  \centering
  \includegraphics[width=12.2cm]{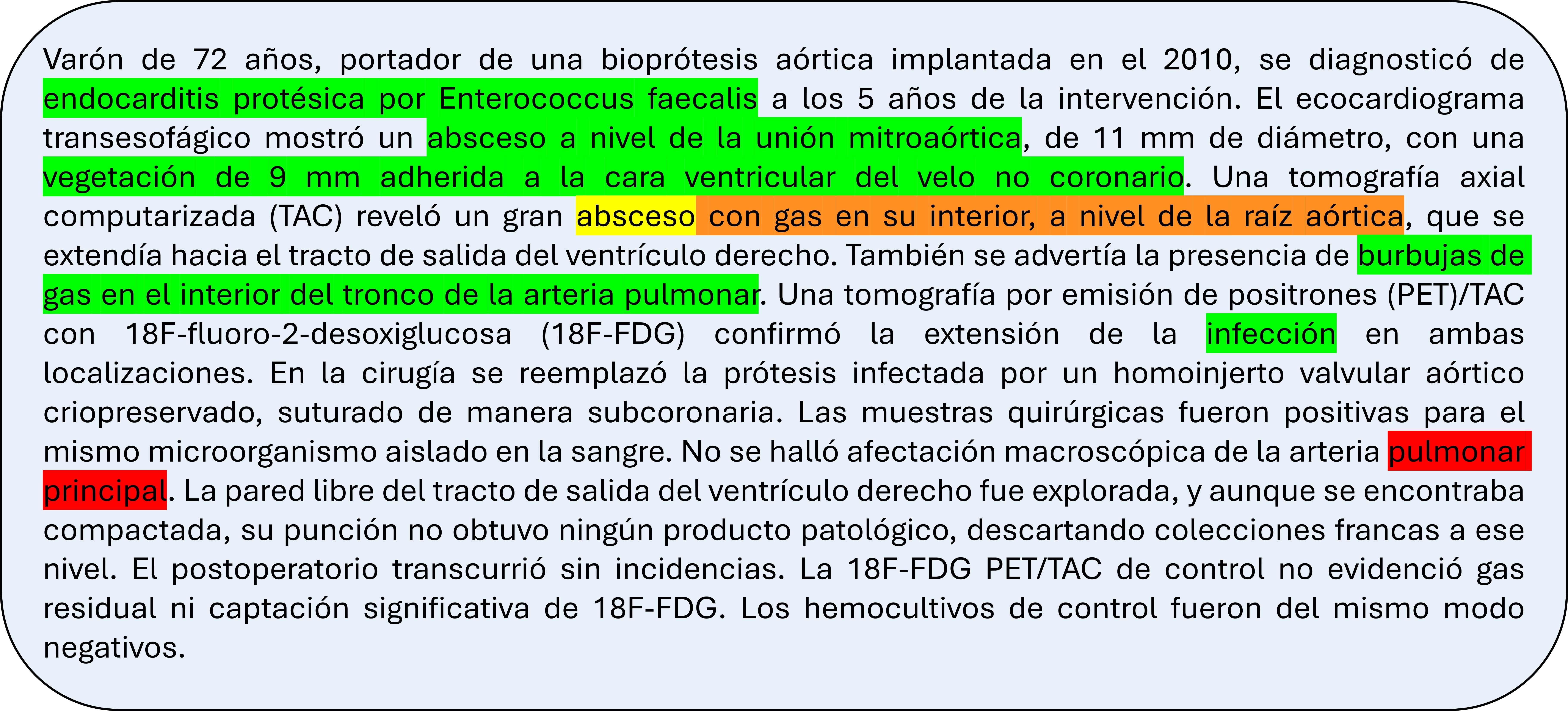}
  \caption{Prediction example for the Spanish Diseases Recognition (SDR) subtask, obtained using the best performing model in terms of F1-score. Green represents correctly identified mentions along with their spans. Red represents mentions that are not present in the ground-truth but predicted by the model. Yellow refers to mentions that are incompletely predicted by the model, while orange marks the full mention as present in the ground-truth.}
  \label{track1_SDR}
\end{figure}

\begin{figure}
  \centering
  \includegraphics[width=12.2cm]{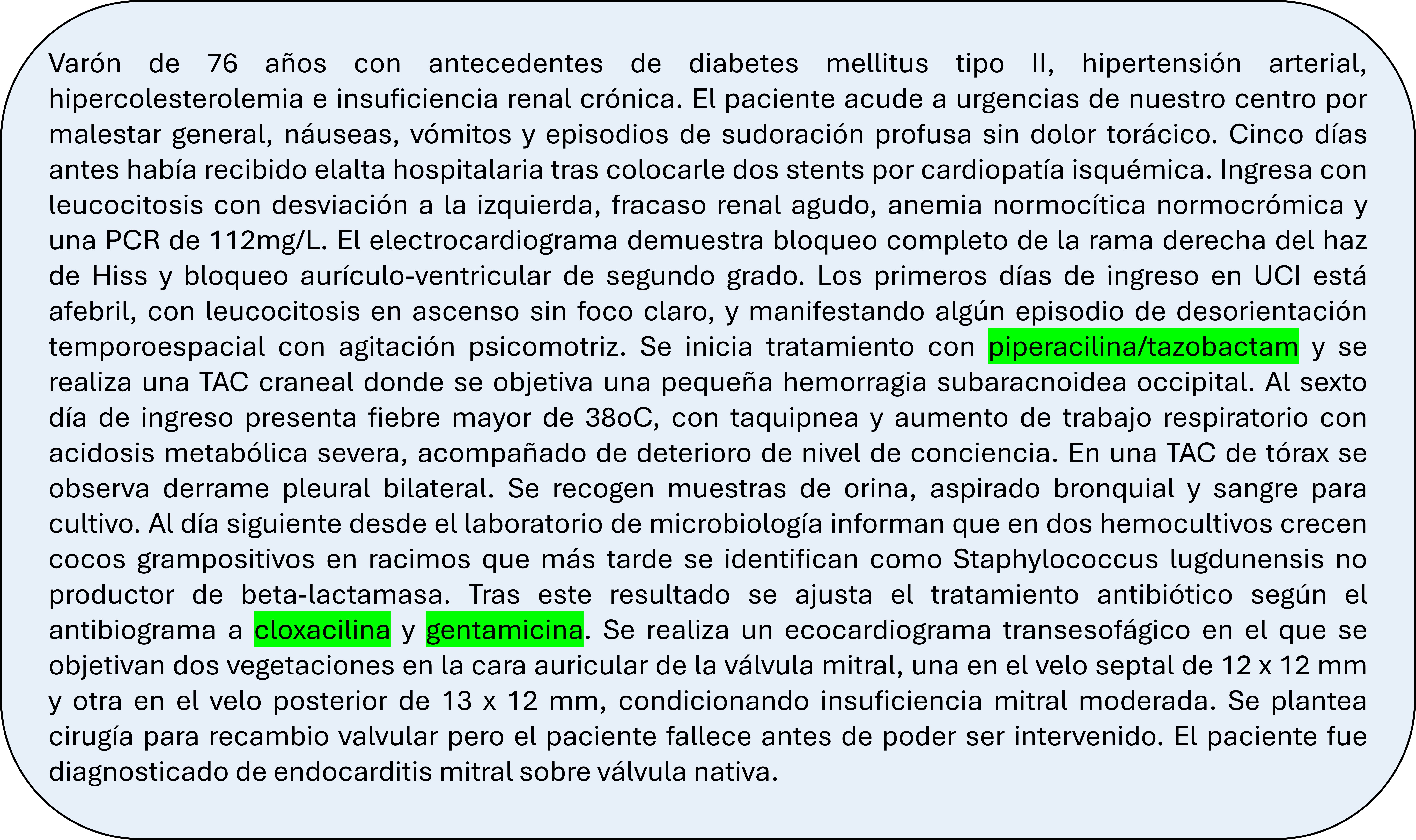}
  \caption{Prediction example for the Spanish Medications Recognition (SMR) subtask, obtained using the best performing model in terms of F1-score. Green represents correctly identified mentions along with their spans. In this particular example, there were no missed, incomplete, or incorrect predictions.}
  \label{track2_SMR}
\end{figure}

\begin{figure}
  \centering
  \includegraphics[width=12.2cm]{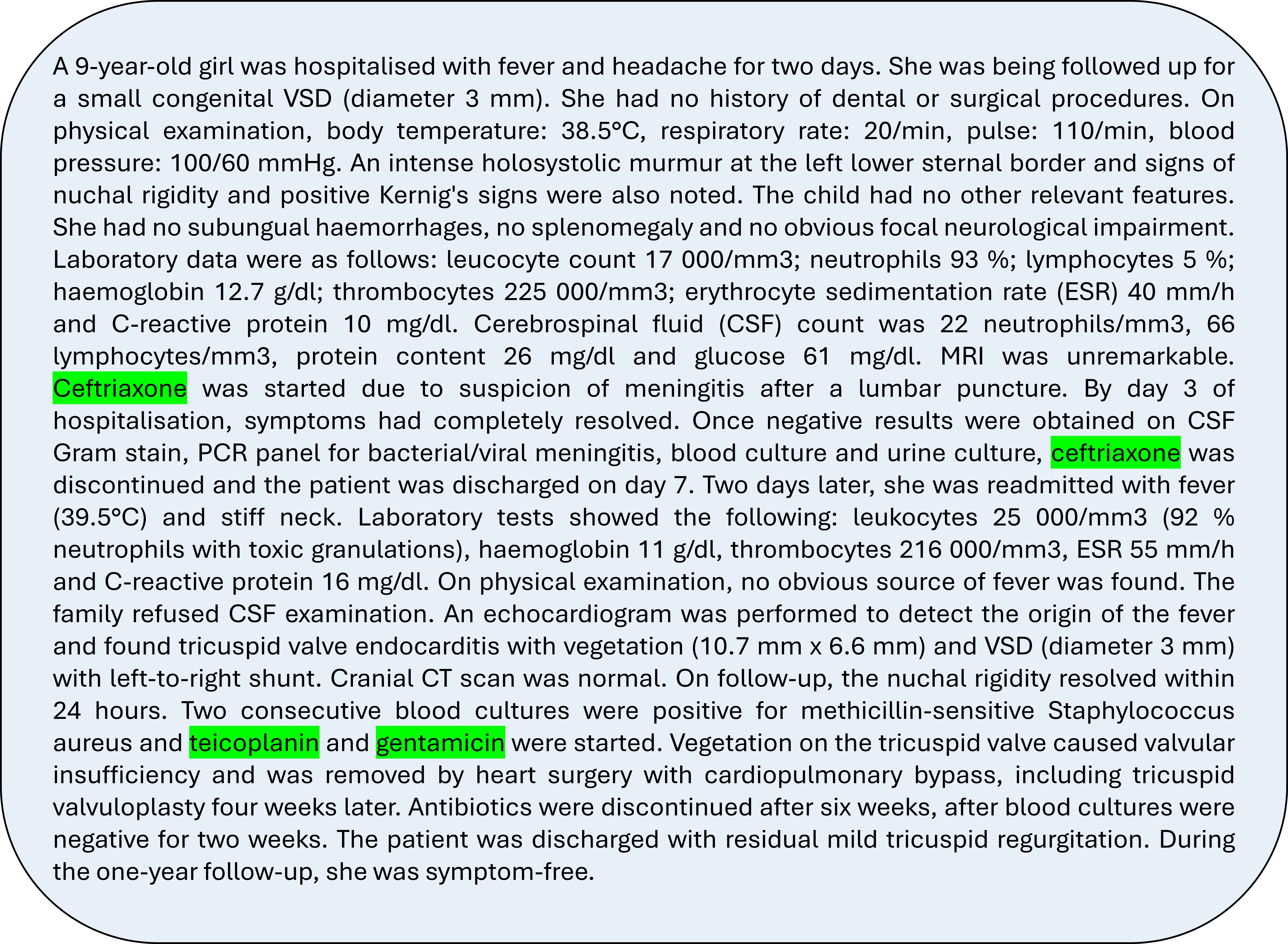}
  \caption{Prediction example for the English Medications Recognition (EMR) subtask, obtained using the best performing model in terms of F1-score. Green represents correctly identified mentions along with their spans. In this particular example, there were no missed, incomplete, or incorrect predictions.}
  \label{track2_EMR}
\end{figure}

\begin{figure}
  \centering
  \includegraphics[width=12.2cm]{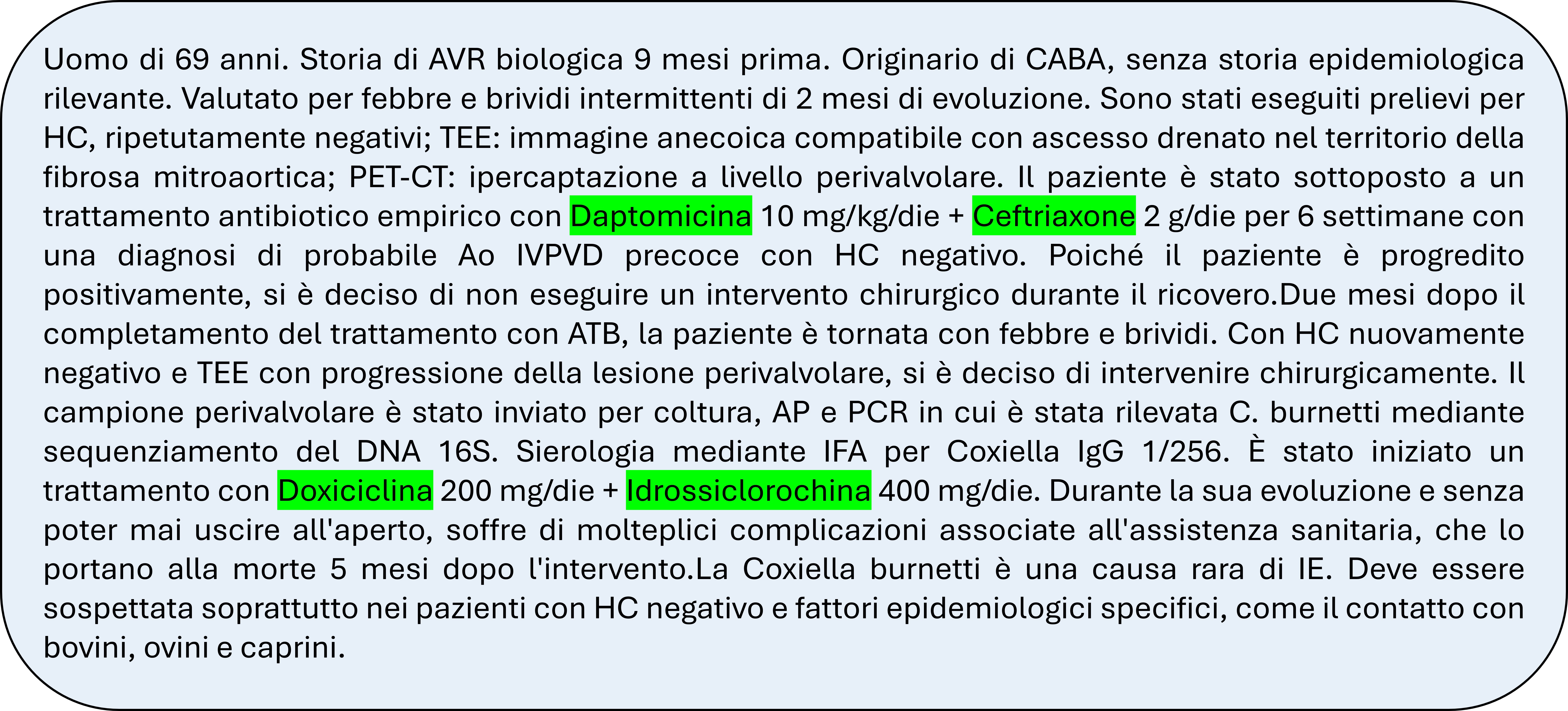}
  \caption{Prediction example for the Italian Medications Recognition (IMR) subtask, obtained using the best performing model in terms of F1-score. Green represents correctly identified mentions along with their spans. In this particular example, there were no missed, incomplete, or incorrect predictions .}
  \label{track2_IMR}
\end{figure}

In addition to this performance analysis, we conducted a qualitative evaluation of the top-performing models across all subtasks. The qualitative analysis complements the quantitative metrics, providing a comprehensive assessment of the capabilities of the developed models in real-world clinical scenarios. The outcomes, as illustrated in Figure~\ref{track1_SDR}, Figure~\ref{track2_SMR}, Figure~\ref{track2_EMR}, and Figure~\ref{track2_IMR}, indicate that the models perform commendably in identifying medications within clinical texts across all three targeted languages. However, the Spanish Diseases Recognition (SDR) model exhibits room for improvement, as it occasionally produces incomplete or incorrect predictions. 

\section{Conclusions}
In this paper, we investigated the utilization of BERT-based contextual embeddings, trained on general domain texts, for extracting mentions of diseases and medications from clinical case reports written in English, Spanish, and Italian. We developed four distinct monolingual models: (1) Spanish Diseases Recognition (SDR),
(2) Spanish Medications Recognition (SMR), (3) English Medications Recognition (EMR), and (4) Italian Medications Recognition (IMR). Additionally, we created two multilingual models: one specialized for Spanish Diseases Recognition (Multi-SDR) and another for Medications Recognition across all three targeted languages (Multi-MMR). While the results show promising performance in identifying medications within clinical texts across all three languages, the models are not flawless. Some weaknesses arise in diseases recognition, where they occasionally produce incomplete or incorrect predictions. To address these issues, we aim to explore the capabilities of recent large language models (LLMs). 

\begin{acknowledgments}
This work received funding from the European Union’s Horizon Europe research and innovation programme
under Grant Agreement No. 101057849 (DataTools4Heart project).
\end{acknowledgments}


\end{document}